\documentclass[lettersize,journal]{IEEEtran}
\usepackage{amsmath,amsfonts}
\usepackage{algorithmic}
\usepackage{algorithm}
\usepackage{array}
\usepackage[caption=false,font=normalsize,labelfont=sf,textfont=sf]{subfig}
\usepackage{textcomp}
\usepackage{stfloats}
\usepackage{url}
\usepackage{verbatim}
\usepackage{graphicx}
\usepackage{cite}
\usepackage{placeins} 
\usepackage{bm}
\usepackage{booktabs}
\usepackage{multirow}
\usepackage{graphicx}   
\usepackage{adjustbox}  
\usepackage{caption}
\usepackage{etoolbox}
\AtBeginEnvironment{table}{\small}

\usepackage{xcolor}
\usepackage[colorlinks=true,
            linkcolor=blue,
            citecolor=red,
            urlcolor=blue]{hyperref}
\usepackage[marginal]{footmisc}

\begin{document}

\title{TimeAPN: Adaptive Amplitude-Phase Non-Stationarity Normalization for Time Series Forecasting}

\author{
Yue Hu,
Jialiang Tang,
Siwei Yu, 
Baosheng Yu, 
Jing Zhang, 
and Dacheng Tao
\thanks{Yue Hu and Siwei Yu are with the School of Mathematics, Harbin Institute of Technology, Harbin 150001, China (e-mail: yuehu@stu.hit.edu.cn; siweiyu@hit.edu.cn). Jialiang Tang is with the School of Computer Science and Engineering, Nanjing University of Science and Technology, China. Baosheng Yu is with the Lee Kong Chian School of Medicine, Nanyang Technological University, Singapore. Jing Zhang is with the School of Computer Science, Wuhan University, Wuhan, China. Dacheng Tao is with the College of Computing and Data Science, Nanyang Technological University, Singapore.}
\thanks{(Yue Hu and Jialiang Tang contributed equally to this work.  Corresponding authors: Siwei Yu and Baosheng Yu.)}%
\thanks{Code is available at \url{https://github.com/y563642-max/timeapn}.}%
}



\maketitle

\begin{abstract}
Non-stationarity is a fundamental challenge in multivariate long-term time series forecasting, often manifested as rapid changes in amplitude and phase. These variations lead to severe distribution shifts and consequently degrade predictive performance. Existing normalization-based methods primarily rely on first- and second-order statistics, implicitly assuming that distributions evolve smoothly and overlooking fine-grained temporal dynamics. To address these limitations, we propose TimeAPN, an Adaptive Amplitude–Phase Non-Stationarity Normalization framework that explicitly models and predicts non-stationary factors from both the time and frequency domains. Specifically, TimeAPN first models the mean sequence jointly in the time and frequency domains, and then forecasts its evolution over future horizons. Meanwhile, phase information is extracted in the frequency domain, and the phase discrepancy between the predicted and ground-truth future sequences is explicitly modeled to capture temporal misalignment. Furthermore, TimeAPN incorporates amplitude information into an adaptive normalization mechanism, enabling the model to effectively account for abrupt fluctuations in signal energy. The predicted non-stationary factors are subsequently integrated with the backbone forecasting outputs through a collaborative de-normalization process to reconstruct the final non-stationary time series. The proposed framework is model-agnostic and can be seamlessly integrated with various forecasting backbones. Extensive experiments on seven real-world multivariate datasets demonstrate that TimeAPN consistently improves long-term forecasting accuracy across multiple prediction horizons and outperforms state-of-the-art reversible normalization methods.

\end{abstract}

\begin{IEEEkeywords}
Time Series Forecasting, Non-Stationary Time Series, Amplitude–Phase Modeling, Frequency-domain Analysis.
\end{IEEEkeywords}

\section{Introduction}

\IEEEPARstart{T}{ime} series forecasting (TSF) aims to predict future temporal dynamics using historical observations, which is a critical task for practical applications ranging from financial forecasting~\cite{clements1999forecasting, granger2014forecasting} and energy management~\cite{alvarez2010energy, deb2017review} to traffic control~\cite{zheng2020gman, yin2021deep} and weather prediction~\cite{bauer2015quiet, ebert2023outlook, allen2025end}. In recent years, deep neural networks (DNNs) have delivered notable gains in TSF owing to their hierarchical representation capabilities, with state-of-the-art performance reported across convolutional neural network (CNN)-based~\cite{wang2023micn, luo2024moderntcn}, Transformer-based~\cite{zhou2021informer, wu2021autoformer, nie2022time, zhu2023time}, and multilayer perceptron (MLP)-based models~\cite{zeng2023transformers, wang2024timemixer}.

However, multivariate time series from complex real-world systems are often non-stationary~\cite{kim2021reversible, liu2023adaptive, mallick2022matchmaker, li2022ddg}. Variations in intrinsic properties, such as period and amplitude, pose significant challenges in linking historical observations with future targets. This issue becomes more pronounced in long-term time series due to the accumulation of forecasting errors, amplification of noise, and the emergence of non-stationary temporal patterns over extended horizons. To further enhance temporal dependency modeling, some approaches incorporate decomposition techniques into DNNs, such as the decomposition of the series into trend-cyclical and seasonal parts~\cite{zeng2023transformers, zhou2022fedformer}, or multiscale components~\cite{wang2024timemixer}. Nevertheless, despite these advances, the performance of DNN-based TSF methods remains unreliable on real-world data, as non-stationarity often leads to discrepancies across temporal horizons and hinders model generalization~\cite{liu2023adaptive}. 

To mitigate the effects of non-stationarity, researchers often normalize the data by removing dynamic factors, typically by scaling each input sequence to have zero mean and unit variance~\cite{passalis2019deep}. The reversible instance normalization (RevIN) method~\cite{kim2021reversible} establishes a general normalization–denormalization framework by introducing a learnable affine transformation, enabling flexible adaptation to diverse temporal data distributions. Building on this idea, subsequent studies have sought to enhance normalization methods to handle the non-stationary problem in TSF. For instance, the slice-level adaptive normalization (SAN) method~\cite{liu2023adaptive} addresses non-stationarity at finer granularity by operating on local temporal slices rather than the entire time series. SAN incorporates a lightweight network module to independently model the statistical properties of the raw data. More recently, the Dual-domain Dynamic Normalization (DDN) method~\cite{dai2024ddn} further performs normalization in both the time and frequency domains by leveraging the wavelet transform to decompose the time series into its constituent frequency components within a sliding window, thereby mitigating non-stationarity.



\begin{figure*}
\centering
\includegraphics[width=0.8\textwidth]{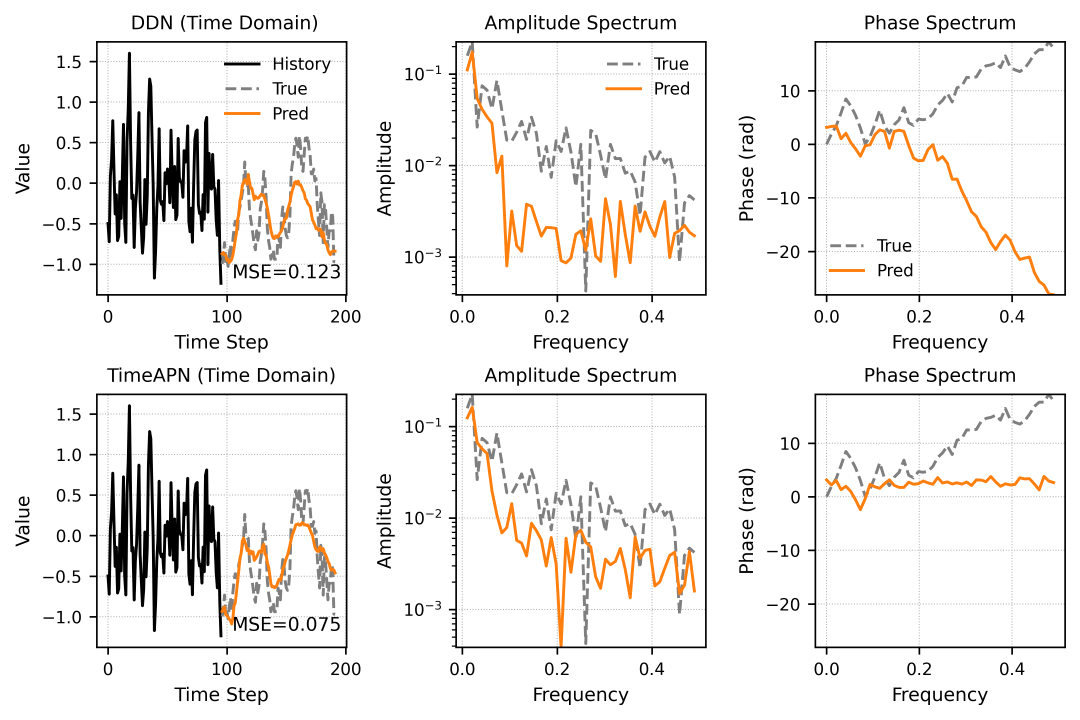}
\caption{Comparison of DDN and TimeAPN in the time and frequency domains. In the time domain (left), TimeAPN achieves a lower prediction error (MSE = 0.075) than DDN (MSE = 0.123). In the frequency domain (middle and right), DDN fails to accurately capture the ground-truth amplitude and phase spectra, whereas TimeAPN produces spectra that align more closely with the true amplitude and phase distributions.}
\label{01_fig}
\end{figure*}

Although existing normalization approaches have improved forecasting performance, they generally overlook the rapid changes in amplitude and phase over time, as shown in Fig.~\ref{01_fig}. These methods primarily focus on first- and second-order statistics, such as the mean and variance, and implicitly simplify the TSF problem by assuming that distributions change smoothly over time. In practice, the amplitude, which reflects the magnitude or strength of the time series, and the phase, which determines its position within a periodic cycle, can change rapidly. Simple statistical measures may only partially capture such dynamic changes, since measures such as the mean and variance describe global tendencies and fail to capture fine-grained time-varying distributions. Therefore, adaptively accounting for these rapid amplitude and phase changes is essential to fully exploit the underlying temporal structure and improve forecasting accuracy. Furthermore, incorporating frequency-domain information provides a complementary perspective for modeling temporal dynamics, enabling the model to better capture rapid variations in amplitude and phase. These observations motivate the development of normalization strategies that dynamically adapt to rapid variations in amplitude and phase.

In this paper, we propose a novel Adaptive Amplitude-Phase Non-stationarity Normalization (TimeAPN) framework that effectively captures time-varying periodic patterns and energy variations in non-stationary time series across both the time and frequency domains. TimeAPN employs the discrete wavelet transform (DWT) to decompose the original time series into multiple frequency components, including both low- and high-frequency parts. Specifically, for each component, the mean sequence is estimated adaptively, and the estimate is then used for energy compensation. To complement the frequency-domain adjustment, we further introduce a time-domain normalization strategy that directly estimates the mean and phase differences from the original time series. Furthermore, TimeAPN explicitly incorporates amplitude information into its adaptive normalization mechanism to capture abrupt fluctuations in signal energy. This adaptive design enables the model to respond effectively to rapid changes in the time series and to produce robust representations for non-stationary time series forecasting.

In summary, the main contributions of this paper are as follows:

\begin{itemize}
    \item We propose a novel plug-in normalization–denormalization framework, termed Adaptive Amplitude–Phase Non-stationarity Normalization (TimeAPN), which explicitly models non-stationary dynamics from both the time and frequency domains. By jointly capturing amplitude variations and phase shifts, TimeAPN enables adaptive normalization that better accommodates rapidly evolving temporal distributions.
    \item We introduce a phase-drift prediction and denormalization correction mechanism that explicitly forecasts future phase variations and incorporates the predicted drift into an invertible denormalization process. This design effectively compensates for temporal misalignment and enhances the robustness of long-horizon forecasting under rapid distribution shifts.
    \item We conduct extensive experiments on seven real-world multivariate benchmarks and four representative forecasting backbones. The results demonstrate that TimeAPN consistently improves long-term forecasting performance and outperforms state-of-the-art reversible normalization methods.
\end{itemize}




The remainder of this paper is organized as follows. Section II reviews related work. Section III introduces the preliminary concepts. Section IV presents the proposed framework in detail. Section V reports the experimental results, discussions, and ablation studies. Finally, Section VI concludes the paper.

\section{RELATED WORK}
\subsection{Time Series Forecasting}
DNN-based methods have become prominent in time series forecasting for their ability to model nonlinear patterns and complex temporal dependencies~\cite{lim2021time, torres2021deep, wang2025mamba}. Architectures such as RNNs~\cite{madan2018predicting, tang2021building}, TCNs~\cite{bai2018empirical, liu2022scinet}, CNNs, MLPs, and Transformers have been widely explored. MICN~\cite{wang2023micn} captures local and global correlations via convolutions, DLinear~\cite{zeng2023transformers} uses a minimalist linear design to model temporal relationships, and TimeMixer~\cite{wang2024timemixer} applies multi-scale decomposition to handle patterns across resolutions. Transformer-based models~\cite{vaswani2017attention, zhou2021informer, nie2022time} further capture long-range dependencies, with PatchTST~\cite{nie2022time} segmenting series into patches and iTransformer~\cite{liu2023itransformer} treating series as tokens to model cross-series correlations. Despite these advances, many DNN-based methods overlook non-stationarity, where evolving dynamics and shifting statistics can degrade forecast accuracy, underscoring the need for approaches that explicitly model changing amplitude and phase characteristics.
\subsection{Non-Stationarity Normalization}
Normalization~\cite{liu2022non, kim2021reversible} stabilizes model training and mitigates time series non-stationarity. Traditional methods, such as z-score normalization~\cite{passalis2019deep}, enforce zero-mean and unit-variance constraints but rely on fixed global statistics, limiting their effectiveness under dynamic conditions. RevIN~\cite{kim2021reversible} introduces a reversible normalization–denormalization scheme with a learnable affine transformation, assuming historical and future sequences share the same distribution. DishTS~\cite{fan2023dish} relaxes this assumption by predicting mean and standard deviation for differing distributions between past and future sequences. SAN~\cite{liu2023adaptive} further captures local distributional changes by estimating adaptive statistics over temporal slices. DDN~\cite{dai2024ddn} incorporates frequency-domain analysis, decomposing time series into components and applying time- and frequency-domain normalization within a sliding window.
Despite these advances, most approaches still rely on mean and variance, assuming smoothly evolving distributions and failing to capture rapid amplitude or phase variations essential for highly non-stationary series.


\subsection{Frequency Analysis}
Frequency-domain analysis effectively reveals periodicity, oscillatory patterns, and signal characteristics across multiple resolutions, with demonstrated applications in time series forecasting (TSF), audio processing, and image analysis~\cite{schroter2022deepfilternet, yi2024filternet, xiao2024frequency, patro2025spectformer}. In TSF, Fourier-transform-based methods extract dominant frequency components and filter high-frequency noise while preserving essential low-frequency information. Wavelet transforms further enable multi-resolution analysis, capturing temporal features from short-term fluctuations to long-term trends~\cite{zhang2025improving}. In audio processing, spectral magnitude analysis supports pitch detection and denoising, while in computer vision, frequency-domain methods facilitate image enhancement, texture analysis, and noise suppression~\cite{gonzalez2002digital, pollock1999time}. Despite their strengths, existing approaches largely emphasize spectral energy, low-frequency trends, or multi-resolution features, often overlooking explicit phase alignment and amplitude calibration~\cite{wang2024timemixer++, huang2025timekan, dai2024ddn}. By contrast, our approach explicitly models both amplitude and phase dynamics, enabling precise adaptation to rapid variations in non-stationary time series.

\section{Preliminaries}

This section introduces the fundamental concepts used in the proposed framework, including the formulation of multivariate time series forecasting, frequency-domain representation via the Fourier Transform, phase compensation in the spectral domain, and the mean-normalization procedure used to extract non-stationary components.

\subsection{Problem Formulation}
In multivariate time series forecasting, a historical sequence $\bm{x} \in \mathbb{R}^{C \times L}$ is given, and the objective is to predict its corresponding future sequence $\bm{y} \in \mathbb{R}^{C \times T}$. Here, $C$ denotes the number of temporal variables, while $L$ and $T$ represent the lengths of the historical (look-back) window and the prediction window, respectively.

\subsection{Frequency Representation}
The Fourier Transform (FT) converts a signal from the time domain to the frequency domain, allowing the identification of dominant periodic components and their associated spectral energy distributions. By incorporating FT-based spectral analysis, the proposed framework provides a comprehensive understanding of both global and local periodic structures and energy variations, thereby enhancing its robustness and adaptability for modeling complex non-stationary time series.

To analyze the frequency characteristics of the time series, the FT $\mathcal{F}$ is applied to the time-domain sequence $\bm{x}^{c'}$, thereby converting it into the frequency domain:
\begin{equation}
\label{eq:fft_def}
\bm{X}^{c'}[k]
= \mathcal{F}\left(\bm{x}^{c'}\right)
= \sum_{n=0}^{L-1} \bm{x}^{c'}[n]\, e^{-j \frac{2\pi}{L}kn}.
\end{equation}
Here, $c' \in \{0, 1, \ldots, C-1\}$ denotes the index of the temporal variable.  $\bm{X}^{c'}[k]$ is obtained by applying the FT to $\bm{x}^{c'}$ and represents the complex number corresponding to the $k^{th}$ frequency component, where $k \in \{0, 1, \ldots, L-1\}$. The complex number at frequency $k$ represents the corresponding frequency component.

In Fourier analysis, a signal can be expressed in the complex frequency domain, where each frequency component is represented by a complex number. This number encodes both the amplitude and phase information, providing a full characterization of the component. The amplitude reflects the magnitude of the component within the original time-domain signal, whereas the phase describes the temporal shift or delay associated with that component. Formally, a frequency component can be represented mathematically as a complex exponential, determined by its specific amplitude and phase:
\begin{equation}\label{eq:02-forwardModel}
	 \bm{X}^{c'}\left[k \right]=\bm{A}^{c'}\left[k \right]e^{j\bm{w}^{c'}\left[k \right]},
\end{equation}
where
\begin{equation*}
    \left\{
    \begin{aligned}
    \bm{A}^{c'}\left[k \right] &= 
    \sqrt{\left(\bm{X}^{c'}_{real}\left[k \right]\right)^{2}
    + \left( \bm{X}^{c'}_{img}\left[k \right]\right)^{2}}, \\[6pt]
    \bm{w}^{c'}\left[k \right] &= 
    \arctan \frac{\bm{X}^{c'}_{img}\left[k \right]}
    {\bm{X}^{c'}_{real}\left[k \right]},
    \end{aligned}
    \right.
\end{equation*}
respectively. Here, $\bm{A}^{c'}\left[k \right]$ represents the amplitude of the $k^{th}$ frequency component, while $\bm{w}^{c'}\left[k \right]$ represents the phase of the $k^{th}$ frequency component. The term $\bm{X}^{c'}_{real}$ denotes the real part of $\bm{X}^{c'}$, and $\bm{X}^{c'}_{img}$ denotes imaginary part of $\bm{X}^{c'}$. This representation fully characterizes the time series in the frequency domain. It enables detailed analysis of both the amplitude and phase properties of its components.

\subsection{Phase Compensation}
Consider another time-domain sequence $\bm{x}^{c''}$. Its frequency-domain representation can similarly be obtained by
\begin{equation}
	 \bm{X}^{c''}=\mathcal{F}(\bm{x}^{c''}),
\end{equation}
where $c'' \in \{0, 1, \ldots, C-1\}$ denotes the index of the temporal variable, and $\bm{X}^{c''}$ is obtained by applying the FT to $\bm{x}^{c''}$. In the frequency domain, this can also be expressed in terms of amplitude and phase:
\begin{equation}
	 \bm{X}{^{c''}}=\bm{A}{^{c''}}e^{j\bm{w}{^{c''}}},
\end{equation}
where $\bm{A}{^{c''}}$ represents the amplitude, and $\bm{w}{^{c''}}$ represents the phase. 

Representing signals in the aforementioned form allows for a direct comparison of amplitude and phase differences between 
$\bm{x}^{c'}$ and $\bm{x}^{c''}$ across all frequencies. In many practical applications, it is desirable to transform $\bm{X}^{c''}$ into $\bm{X}^{c'}$ through amplitude and phase compensation. This relationship can be formulated as:
\begin{equation}\label{eq:05-forwardModel}
	 \bm{X}^{c'}=\frac{ \bm{A}^{c'}}{\bm{A}^{c''}}e^{j \Delta\bm{w}}\bm{X}^{c''},
\end{equation}
where $\frac{ \bm{A}^{c'}}{\bm{A}^{c''}}$ denotes the amplitude compensation factor and adjusts the amplitude discrepancy between $\bm{X}^{c''}$ and $\bm{X}^{c'}$, and $\Delta\bm{w}=\bm{w}^{c'}-\bm{w}^{c''}$ represents the phase compensation factor and corrects any phase difference. By applying these compensation operators, the frequency components of $\bm{X}^{c''}$ can be aligned with those of $\bm{X}^{c'}$, enabling accurate frequency-domain matching or correction.

To reconstruct the time series in the time domain, the Inverse Fourier Transform (IFT) operator $\mathcal{F}^{-1}$ is applied to both sides of the equation \eqref{eq:02-forwardModel}. The IFT, defined as
\begin{equation}
\label{eq:ifft_def}
\bm{x}^{c'}[n]
= \mathcal{F}^{-1}(\bm{X}^{c'})
= \frac{1}{L}
\sum_{k=0}^{L-1}
\bm{X}^{c'}[k]\, 
e^{j \frac{2\pi}{L} kn}.
\end{equation}
where $n \in \{0, 1, \ldots, L-1\}$. Eq.~\eqref{eq:ifft_def} converts the frequency-domain representation back into its corresponding form in the time domain. Multiplication by the compensation function in the frequency domain is equivalent to convolution with its inverse Fourier transform in the time domain. Accordingly, we obtain:
\begin{equation}\label{eq:07-forwardModel}
	 \bm{x}^{{c'}}=\mathcal{F}^{-1} \left(\frac{ \bm{A}^{c'}}{\bm{A}^{c''}}e^{j \Delta\bm{w}} \right)*\bm{x}^{c''},
\end{equation}
where $*$ denotes the convolution operator in the time domain. This equation indicates that the time series $\bm{x}^{c'}$ can be derived by convolving $\bm{x}^{c''}$ with the IFT of the complex compensation term $\frac{ \bm{A}^{c'}}{\bm{A}^{c''}}e^{j \Delta\bm{w}}$.

\subsection{Mean Normalization}
Following the SAN method \cite{liu2023adaptive}, TimeAPN computes the point-wise mean value of the time series over a centered sliding window. Concretely, for each channel $c$ and each time index $t$, the mean value of ${\bm{x}}_{t}^{c}$ is computed over the window $[t-s, t+s]$: 
\begin{equation}\label{eq:08-forwardModel}
\mu_{t}^{c} = \frac{1}{2s+1} \sum_{h=-s}^{s} {\bm{x}}_{h+t}^{c}.
\end{equation}
Here, the parameter $2s + 1$ denotes the size of the sliding window, with a stride of 1, and $\mu_t^c\in\mathbb{R}$ represents the mean of channel $c$ at the $t^{th}$ time point, where $t \in \{s + 1, \ldots, L - s\}$. $c \in \{0, 1, \ldots, C-1\}$ denotes the index of the temporal variable. To ensure that each time point has a corresponding mean value, the values near the sequence boundaries are extended using the $\mathrm{ReplicationPad} (\cdot)$ operation. This padding copies the nearest available mean values in time so that the resulting sequence of per-window mean values is temporally aligned with the original input length:
\begin{equation}\label{eq:09-forwardModel}
\bm{\mu}^{c} = \mathrm{ReplicationPad} \left( \mu_{s+1}^{c},\ldots, \mu_{L-s}^{c} \right).
\end{equation}
Later, we normalize the original input sequence by its individual mean value as:
\begin{equation}\label{eq:10-forwardModel}
\bar{\bm{x}}^{c} = \bm{x}^{c} - \bm{\mu}^{c},
\end{equation}
where $\bm{x}^c$ is the historical time series of the $c^{th}$ channel.

We transform the normalized input sequence $\bar{\bm{x}}^c$ into frequency using $\mathcal{F}$, and represent it in the same form as Eq.~\eqref{eq:02-forwardModel}. The amplitude and phase of $\bar{\bm{x}}^c$ are then computed as:
\begin{equation}\label{eq:11-forwardModel}
    \begin{aligned}
    \bar{\bm{A}}^{c} &= 
    \sqrt{\left(\bar{\bm{X}}^{c}_{real}\right)^{2}
    + \left( \bar{\bm{X}}^{c}_{img}\right)^{2}}, \\[6pt]
    \bar{\bm{w}}^{c} &= 
    \arctan \frac{\bar{\bm{X}}^{c}_{img}}
    {\bar{\bm{X}}^{c}_{real}},
    \end{aligned}
\end{equation}
where $\bar{\bm{A}}^c$ represents the amplitude, and $\bar{\bm{w}}^c$ represents the phase. 
Through the procedures described in equations \eqref{eq:08-forwardModel}-\eqref{eq:11-forwardModel}, we define the operation
\begin{equation}\label{eq:12-forwardModel}
\mathrm{MeanNormPhase}(\cdot):\, \bm{x}^{c} \;\longmapsto\; \left( \bar{\bm{x}}^{c}, \bm{\mu}^{c}, \, \bar{\bm{w}}^{c} \right),
\end{equation}
which takes the original input sequence $\bm{x}^{c}$, computes its sliding-window mean sequence $\bm{\mu}^{c}$, and obtains the mean-normalized sequence $\bar{\bm{x}}^{c} =\bm{x}^{c}-\bm{\mu}^{c}$. The operation then transforms $\bar{\bm{x}}^{c}$ into the frequency domain and outputs the phase $\bar{\bm{w}}^{c}$ of the mean-normalized sequence. In this way, the non-stationary factors at each time point can be effectively extracted.

\begin{figure*}[!ht]
\centering
\includegraphics[width=1.00\textwidth]{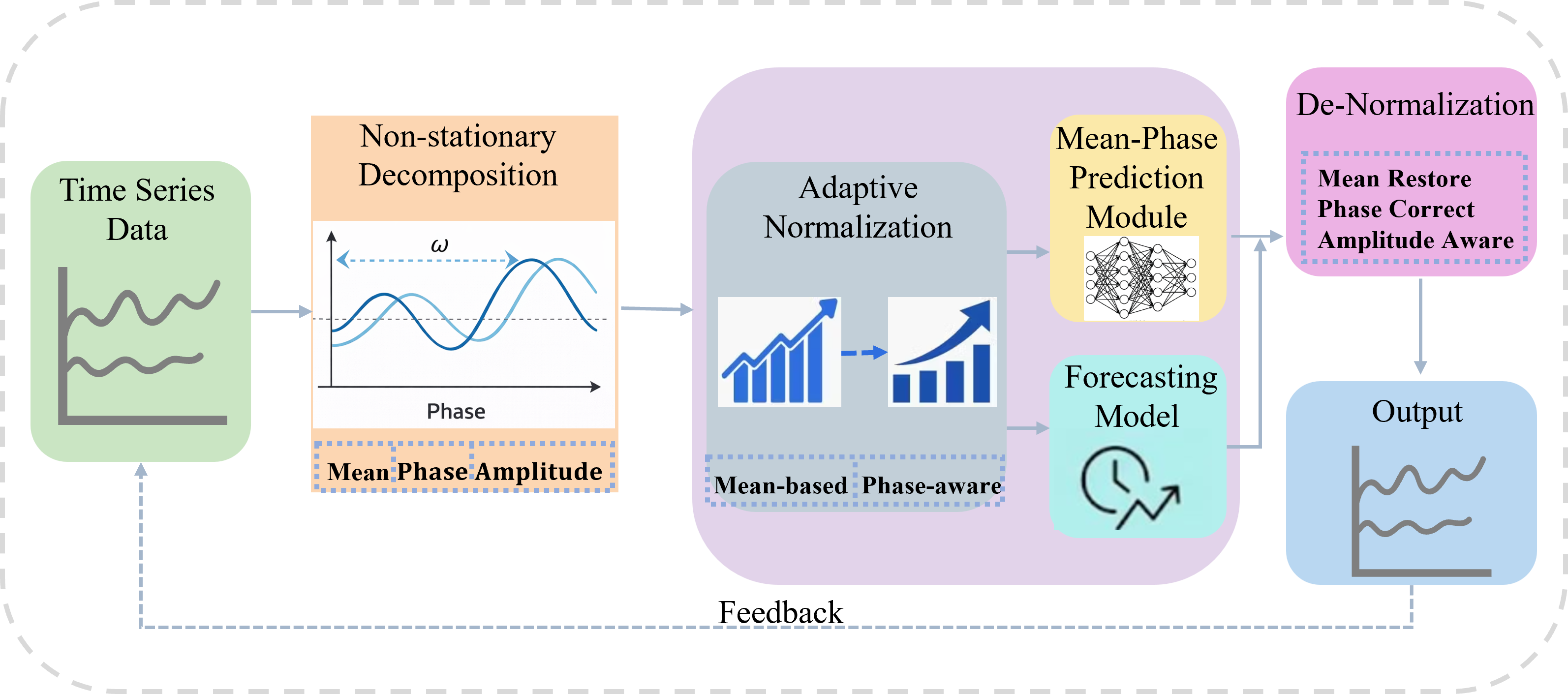}
\caption{Overview of the proposed TimeAPN framework. 
The model first normalizes the input sequence to extract stationary components and non-stationary factors. Stationary sequences are fed into a forecasting model (FM), while non-stationary factors are predicted using the Mean and Phase Prediction Module (MPPM). Predicted non-stationary factors are then combined with FM outputs through de-normalization to reconstruct the future series.
}
\label{02_fig}
\end{figure*}

\section{Method}
 In this section, we present the TimeAPN framework and detail its normalization, non-stationarity prediction, de-normalization, and collaborative training mechanisms.

\subsection{Overview}

As illustrated in Fig. \ref{02_fig}, the proposed framework first performs normalization that incorporates both phase and mean-value information, thereby mitigating non-stationarity and improving the accuracy of future predictions. This process produces two stationary sequences alongside two corresponding sets of non-stationary factors. The stationary sequences are adaptively weighted and input to the time-series Forecasting Model (FM) to predict future sequences, while the non-stationary factors are processed by the Mean Value and Phase Prediction Model (MPPM) to estimate future mean and phase variations. The predicted non-stationary factors are then reweighted and integrated with the FM outputs through a de-normalization procedure to reconstruct the final non-stationary time series. All components are trained collaboratively, leveraging mean, amplitude, and phase information to ensure coherent learning and improve overall forecasting accuracy.


\subsection{Normalization}
The proposed TimeAPN framework exploits local mean, amplitude, and phase information to characterize fast-changing non-stationary dynamics, including abrupt fluctuations in signal energy and phase alignment, which commonly arise in non-stationary time series. The framework explicitly models the mean and phase components to isolate fast-varying dynamics from the original series. Furthermore, the optimization objective incorporates amplitude information, which allows rapid changes in energy to be captured and modeled accurately.

\paragraph{Frequency-Domain Normalization} We analyze the signal in the frequency domain to identify and precisely remove the non-stationary component. The DWT decomposes the time series into a low-frequency component $\bm{x}_l^{c}$ and a high-frequency component $\bm{x}_h^{c}$. The low-frequency component primarily reflects the overall trend, whereas the high-frequency component captures transient variations. This frequency-domain decomposition facilitates the identification of a broader spectrum of non-stationary factors that are difficult to distinguish in the time domain. By independently analyzing and processing these components, non-stationary elements can be accurately identified and effectively removed, thereby improving the stability and discriminability of the time series in the frequency domain. The non-stationary factor within these components is subsequently extracted and eliminated through further processing. The overall procedure is formulated as follows:
\begin{equation}
\label{eq:idwt_phase}
\begin{aligned}
	\bm{x}_h^{c},\bm{x}_l^{c} &= \mathrm{DWT}_{\psi_{l,h}}\left( \bm{x}^c\right),\\
    \bar{\bm{x}}^c_h, \bm{\mu}^{c}_h, \sim &= \mathrm{MeanNormPhase}\left( \bm{x}_h^{c}\right),\\
    \bar{\bm{x}}^c_l, \bm{\mu}^{c}_l, \sim &= \mathrm{MeanNormPhase}\left( \bm{x}_l^{c}\right),  
\end{aligned}
\end{equation}
Here, $\psi_{l,h}$ is a pair of learnable wavelet bases. $\bar{\bm{x}}^c_h, \bm{\mu}^{c}_h$ denote, respectively, the stationary sequence and mean of the high-frequency component. $\bar{\bm{x}}^c_l, \bm{\mu}^{c}_l$ denote those of the low-frequency component.

In practice, different implementations of the DWT may adopt distinct padding schemes, leading to variations in the lengths of the resulting output sequences. To ensure consistency in output size, the Inverse Discrete Wavelet Transform (IDWT) is subsequently applied to reconstruct the sequences to a predefined length. Beyond aligning sequence lengths, the IDWT also restores the signal in the time domain, thereby reducing reconstruction artifacts and suppressing high-frequency noise components introduced during the DWT process. These procedures can be formulated as follows:
\begin{equation}
\label{eq:idwt_group}
\begin{aligned}
	\bar{\bm{x}}^{c}_\mathrm{f} &= \mathrm{IDWT}_{\psi_{l,h}}\left(\bar{\bm{x}}^c_h, \bar{\bm{x}}^c_l\right),\\
    \bm{\mu}^{c}_\mathrm{f} &= \mathrm{IDWT}_{\psi_{l,h}}\left(\bm{\mu}^c_h, \bm{\mu}^c_l\right),
\end{aligned}
\end{equation}
where $\bar{\bm{x}}^{c}_\mathrm{f}$ and $\bm{\mu}^{c}_\mathrm{f}$ denote the stationary and mean sequences respectively.

\paragraph{Time-Domain Normalization} Normalization is also performed directly in the time domain to capture fine-grained temporal variations. The time-domain stationary sequence, mean, and phase are obtained as:
\begin{equation}\label{eq:19-forwardModel}
	\bar{\bm{x}}^c_\mathrm{t}, \bm{\mu}^{c}_\mathrm{t}, \bar{\bm{w}}^c_\mathrm{t}= \mathrm{MeanNormPhase}\left( \bm{x}^{c}\right).
\end{equation}

\paragraph{Fusion} The time-domain and frequency-domain stationary sequences are then fused through a weighted operation, defined as:
\begin{equation}\label{eq:20-forwardModel}
	\bar{\bm{x}}^c = \alpha\bar{\bm{x}}^c_\mathrm{t} + \left( 1-\alpha\right)\bar{\bm{x}}^c_\mathrm{f},
\end{equation}
where $\alpha$ is a learnable parameter. The fused sequence $\bar{\bm{x}}^c$ serves as the final sequence and is then input to FM for forecasting. 
$\bm{\mu}^c_\mathrm{t}, \bm{\mu}^c_\mathrm{f}$ are used to address the problem of inaccurate energy distribution in the FM. The loss function incorporates amplitude information, in addition to mean normalization, to accurately capture rapid fluctuations in signal energy and improve modeling performance. $\bar{\bm{w}}^c_\mathrm{t}$ is used as the phase term to mitigate prediction bias caused by phase drift in the FM. The corresponding mean and phase information at each time point are then provided as inputs to the MPPM. 

\subsection{Non-Stationarity Prediction}
Following the procedures introduced in the previous subsection, we extract the mean from both the time and frequency domains and the phase component from the frequency domain. The model then explicitly predicts future mean values and phase differences to capture the rapidly changing components in the time series. This non-stationarity prediction mechanism allows the model to dynamically adapt to abrupt distributional changes and provide a precise representation of temporal dynamics. 
We compute the mean of the extracted mean sequence to obtain the difference sequence. The difference sequence, together with the original time series and the corresponding phase series, is used to predict future mean values. The predicted mean values are then added back to their respective mean value to obtain the final mean output. We employ an MLP for mean forecasting. The overall procedure is formulated as follows:
\begin{equation}\label{eq:21-forwardModel}
\begin{aligned}
\Delta \bm{\mu}^c_\mathrm{f'} &= \mathrm{MLP}\left( \bm{\mu}^c_\mathrm{f} - a^c, \bm{x}^c \right), \quad
\Delta \bm{\mu}^c_\mathrm{f} &= \Delta \bm{\mu}^c_\mathrm{f'} + a^c.
\end{aligned}
\end{equation}
Here, $a^c$ is the mean value of $\bm{\mu}^c_\mathrm{f}$, while $\Delta \bm{\mu}^c_\mathrm{f}$ denotes the predicted mean sequence in the frequency domain. The $\mathrm{MLP}$ is a mean prediction branch. 
As the above frequency domain prediction process, for the time domain, this prediction process can be formulated as follows:
\begin{equation}\label{eq:forwardModel_t}
\begin{aligned}
\Delta \bm{\mu}^c_\mathrm{t'} &= \mathrm{MLP}\left( \bm{\mu}^c_\mathrm{t} - b^c, \bm{x}^c \right), \quad
\Delta \bm{\mu}^c_\mathrm{t} = \Delta \bm{\mu}^c_\mathrm{t'} + b^c, \\
\Delta \bm{w}^c_\mathrm{t} &= \mathrm{TCN}\left( \bar{\bm{w}}^c_\mathrm{t} \right).
\end{aligned}
\end{equation}
Here, $b^c$ is the mean values of $\bm{\mu}^c_\mathrm{t}$, while $\Delta\bm{\mu}^c_\mathrm{t}$ and $\Delta{\bm{w}}^c_\mathrm{t}$ denote the predicted mean sequence and phase difference in the time domain, respectively. The $\mathrm{TCN}$ is a phase prediction branch. 

\subsection{De-normalization} 
With the prediction of mean and phase difference, TimeAPN feeds the stationary sequence into the FM, which produces an intermediate representation denoted as ${\bar{\bm{y}}}^c$. To obtain the final prediction, TimeAPN first applies a weighted reconstruction to restore the non-stationary factors and then performs de-normalization on the backbone’s output. The overall procedure can be expressed as follows:
\begin{equation}
\label{eq:23-forwardModel}
\begin{aligned}
    \Delta\bm{\mu}^c &= \alpha\Delta\bm{\mu}^c_\mathrm{t} + (1-\alpha)\Delta\bm{\mu}^c_\mathrm{f}, \quad 
    \Delta\bar{\bm{w}}^c = \Delta{\bm{w}}^c_\mathrm{t},\\
    \bm{y}^c_* &= \mathcal{F}^{-1}\!\left(e^{j \Delta\bar{\bm{w}}^c} \right)*\bar{\bm{y}}^c + \Delta{\bm{\mu}}^c,
\end{aligned}
\end{equation}
where $\bm{y}^c_*$ represents the predicted future sequence after incorporating non-stationary information, and $\bar{\bm{y}}^c$ is the predicted future series of FM. $\bm{y}^c_*$ will be later used for loss computation ($\mathcal{L}_{\mathrm{FM}}$) and performance evaluation. To simplify the problem, we did not adopt the reconstruction method for non-stationary information described in Eq.~\eqref{eq:07-forwardModel}. Instead, the amplitude information was directly added to the loss computation.

\begin{table*}[ht]
\centering
\small
\caption{
Multivariate long-term forecasting results on benchmark datasets under prediction lengths of 96, 192, 336, and 720. Results are reported in terms of MSE and MAE. TimeAPN is integrated into four representative backbone models (FEDformer, DLinear, PatchTST, and S\textunderscore Mamba). The best results are highlighted in \textbf{bold}.
}
\renewcommand{\arraystretch}{1.1}

\resizebox{\textwidth}{!}{%
\begin{tabular}{cc|cccc|cccc|cccc|cccc}
\toprule
\multicolumn{2}{c}{{Methods}} &
\multicolumn{2}{c}{FEDformer} &
\multicolumn{2}{c}{+TimeAPN} &
\multicolumn{2}{c}{DLinear} &
\multicolumn{2}{c}{+TimeAPN} &
\multicolumn{2}{c}{PatchTST} &
\multicolumn{2}{c}{+TimeAPN} &
\multicolumn{2}{c}{S\textunderscore Mamba} &
\multicolumn{2}{c}{+TimeAPN} \\
\cmidrule(lr){3-4}\cmidrule(lr){5-6}\cmidrule(lr){7-8}\cmidrule(lr){9-10}\cmidrule(lr){11-12}\cmidrule(lr){13-14}\cmidrule(lr){15-16}\cmidrule(lr){17-18}
\multicolumn{2}{c}{Metric}& MSE & MAE & MSE & MAE & MSE & MAE & MSE & MAE & MSE & MAE & MSE & MAE & MSE & MAE & MSE & MAE \\
\bottomrule

\multirow{4}{*}{\rotatebox{90}{ETTh1}}
& 96  & \textbf{0.375} & 0.415 & 0.377 & \textbf{0.399} & 0.384 & 0.405 & \textbf{0.371} & \textbf{0.393} & 0.393 & 0.408 & \textbf{0.382} & \textbf{0.392} & 0.388 & 0.406 & \textbf{0.382} & \textbf{0.396} \\
& 192 & 0.427 & 0.448 & \textbf{0.424} & \textbf{0.435} & 0.443 & 0.450 & \textbf{0.405} & \textbf{0.415} & 0.445 & 0.434 & \textbf{0.430} & \textbf{0.420} & 0.445 & 0.441 & \textbf{0.439} & \textbf{0.427} \\
& 336 & \textbf{0.458} & 0.465 & 0.466 & \textbf{0.458} & 0.447 & 0.448 & \textbf{0.430} & \textbf{0.431} & 0.484 & 0.451 & \textbf{0.472} & \textbf{0.442} & 0.490 & 0.465 & \textbf{0.484} & \textbf{0.451} \\
& 720 & 1.093 & 0.827 & \textbf{0.498} & \textbf{0.499} & 0.504 & 0.515 & \textbf{0.435} & \textbf{0.454} & 0.480 & 0.471 & \textbf{0.468} & \textbf{0.464} & 0.506 & 0.497 & \textbf{0.481} & \textbf{0.473} \\
\midrule

\multirow{4}{*}{\rotatebox{90}{ETTh2}} 
& 96  & 0.341 & 0.385 & \textbf{0.288} & \textbf{0.343} & 0.290 & 0.353 & \textbf{0.272} & \textbf{0.335} & 0.298 & 0.347 & \textbf{0.289} & \textbf{0.338} & 0.297 & 0.349 & \textbf{0.290} & \textbf{0.340} \\
& 192 & 0.433 & 0.441 & \textbf{0.377} & \textbf{0.404} & 0.388 & 0.422 & \textbf{0.334} & \textbf{0.376} & 0.381 & 0.396 & \textbf{0.369} & \textbf{0.390} & 0.378 & 0.399 & \textbf{0.372} & \textbf{0.394} \\
& 336 & 0.503 & 0.495 & \textbf{0.432} & \textbf{0.444} & 0.463 & 0.473 & \textbf{0.360} & \textbf{0.400} & \textbf{0.383} & \textbf{0.411} & 0.409 & 0.422 & 0.425 & 0.435 & \textbf{0.413} & \textbf{0.430} \\
& 720 & 0.480 & 0.485 & \textbf{0.439} & \textbf{0.454} & 0.733 & 0.606 & \textbf{0.398} & \textbf{0.436} & 0.412 & 0.434 & \textbf{0.402} & \textbf{0.431} & 0.432 & 0.448 & \textbf{0.420} & \textbf{0.443} \\
\midrule

\multirow{4}{*}{\rotatebox{90}{ETTm1}}
& 96  & 0.363 & 0.413 & \textbf{0.308} & \textbf{0.364} & 0.301 & 0.345 & \textbf{0.287} & \textbf{0.341} & 0.321 & 0.360 & \textbf{0.316} & \textbf{0.354} & 0.331 & 0.368 & \textbf{0.320} & \textbf{0.366} \\
& 192 & 0.399 & 0.429 & \textbf{0.350} & \textbf{0.391} & 0.336 & 0.366 & \textbf{0.325} & \textbf{0.364} & 0.365 & 0.382 & \textbf{0.355} & \textbf{0.381} & 0.378 & 0.393 & \textbf{0.372} & \textbf{0.391} \\
& 336 & 0.440 & 0.456 & \textbf{0.391} & \textbf{0.420} & 0.372 & 0.389 & \textbf{0.358} & \textbf{0.384} & 0.392 & \textbf{0.404} & \textbf{0.385} & \textbf{0.404} & 0.410 & 0.414 & \textbf{0.403} & \textbf{0.412} \\
& 720 & 0.492 & 0.480 & \textbf{0.443} & \textbf{0.449} & 0.427 & 0.423 & \textbf{0.413} & \textbf{0.417} & \textbf{0.451} & \textbf{0.434} & \textbf{0.451} & 0.439 & 0.474 & 0.451 & \textbf{0.468} & \textbf{0.448} \\
\midrule

\multirow{4}{*}{\rotatebox{90}{ETTm2}}
& 96  & 0.190 & 0.283 & \textbf{0.168} & \textbf{0.256} & 0.172 & 0.267 & \textbf{0.168} & \textbf{0.261} & 0.176 & 0.260 & \textbf{0.175} & \textbf{0.257} & 0.182 & 0.266 & \textbf{0.174} & \textbf{0.259} \\
& 192 & 0.256 & 0.324 & \textbf{0.236} & \textbf{0.304} & 0.237 & 0.314 & \textbf{0.223} & \textbf{0.298} & 0.244 & 0.304 & \textbf{0.241} & \textbf{0.301} & 0.252 & 0.313 & \textbf{0.241} & \textbf{0.304} \\
& 336 & 0.326 & 0.365 & \textbf{0.302} & \textbf{0.343} & 0.295 & 0.359 & \textbf{0.275} & \textbf{0.332} & 0.310 & 0.346 & \textbf{0.301} & \textbf{0.340} & 0.313 & 0.349 & \textbf{0.301} & \textbf{0.341} \\
& 720 & 0.439 & 0.428 & \textbf{0.400} & \textbf{0.410} & 0.427 & 0.439 & \textbf{0.363} & \textbf{0.389} & 0.402 & 0.400 & \textbf{0.397} & \textbf{0.397} & 0.413 & 0.405 & \textbf{0.396} & \textbf{0.401} \\
\midrule

\multirow{4}{*}{\rotatebox{90}{Weather}}
& 96  & 0.240 & 0.326 & \textbf{0.173} & \textbf{0.249} & 0.174 & 0.233 & \textbf{0.146} & \textbf{0.203} & 0.183 & 0.222 & \textbf{0.167} & \textbf{0.219} & 0.165 & \textbf{0.210} & \textbf{0.155} & 0.220 \\
& 192 & 0.449 & 0.464 & \textbf{0.228} & \textbf{0.287} & 0.218 & 0.278 & \textbf{0.190} & \textbf{0.249} & 0.229 & \textbf{0.261} & \textbf{0.214} & 0.264 & 0.214 & \textbf{0.252} & \textbf{0.205} & 0.264 \\
& 336 & 0.358 & 0.398 & \textbf{0.285} & \textbf{0.329} & 0.263 & 0.314 & \textbf{0.239} & \textbf{0.289} & 0.283 & \textbf{0.300} & \textbf{0.266} & 0.305 & 0.274 & \textbf{0.297} & \textbf{0.254} & 0.303 \\
& 720 & 0.411 & 0.425 & \textbf{0.356} & \textbf{0.404} & 0.332 & 0.374 & \textbf{0.309} & \textbf{0.340} & 0.355 & \textbf{0.347} & \textbf{0.332} & 0.358 & 0.350 & \textbf{0.345} & \textbf{0.332} & 0.360 \\
\midrule

\multirow{4}{*}{\rotatebox{90}{Electricity}}
& 96  & 0.188 & 0.304 & \textbf{0.144} & \textbf{0.244} & 0.140 & 0.237 & \textbf{0.128} & \textbf{0.225} & 0.186 & 0.268 & \textbf{0.131} & \textbf{0.228} & 0.137 & \textbf{0.235} & \textbf{0.135} & 0.240 \\
& 192 & 0.197 & 0.311 & \textbf{0.162} & \textbf{0.260} & 0.154 & 0.250 & \textbf{0.147} & \textbf{0.245} & 0.189 & 0.273 & \textbf{0.148} & \textbf{0.246} & 0.161 & \textbf{0.259} & \textbf{0.154} & \textbf{0.259} \\
& 336 & 0.213 & 0.327 & \textbf{0.177} & \textbf{0.283} & 0.169 & 0.268 & \textbf{0.160} & \textbf{0.261} & 0.205 & 0.288 & \textbf{0.164} & \textbf{0.264} & 0.179 & \textbf{0.276} & \textbf{0.170} & \textbf{0.276} \\
& 720 & 0.242 & 0.353 & \textbf{0.198} & \textbf{0.310} & 0.204 & 0.300 & \textbf{0.187} & \textbf{0.288} & 0.246 & 0.321 & \textbf{0.201} & \textbf{0.299} & 0.207 & \textbf{0.301} & \textbf{0.196} & 0.304 \\
\midrule

\multirow{4}{*}{\rotatebox{90}{Traffic}}
& 96  & 0.575 & 0.357 & \textbf{0.449} & \textbf{0.283} & 0.413 & 0.287 & \textbf{0.374} & \textbf{0.262} & 0.372 & 0.259 & \textbf{0.362} & \textbf{0.258} & \textbf{0.353} & \textbf{0.253} & 0.360 & 0.254 \\
& 192 & 0.613 & 0.381 & \textbf{0.470} & \textbf{0.298} & 0.424 & 0.290 & \textbf{0.393} & \textbf{0.272} & 0.384 & \textbf{0.263} & \textbf{0.375} & 0.265 & \textbf{0.354} & \textbf{0.261} & 0.370 & 0.264 \\
& 336 & 0.621 & 0.380 & \textbf{0.476} & \textbf{0.316} & 0.438 & 0.299 & \textbf{0.409} & \textbf{0.280} & 0.398 & \textbf{0.270} & \textbf{0.395} & 0.278 & \textbf{0.368} & \textbf{0.264} & 0.390 & 0.278 \\
& 720 & 0.630 & 0.382 & \textbf{0.520} & \textbf{0.320} & 0.466 & 0.316 & \textbf{0.447} & \textbf{0.300} & 0.435 & \textbf{0.290} & \textbf{0.432} & 0.298 & \textbf{0.420} & \textbf{0.288} & 0.438 & 0.302 \\
\bottomrule
\end{tabular}%
} 
\label{tab:longterm}
\end{table*}

\subsection{Collaborative Training}

TimeAPN formulates the learning process as a bi-level optimization problem, in which the lower-level objective captures non-stationary dynamics, and the upper-level objective enhances forecasting performance. To address the complexity of this formulation, TimeAPN employs a two-stage training strategy. In the first stage, the Mean and Phase Prediction Module (MPPM) is trained to minimize amplitude and phase discrepancies, allowing the model to extract intrinsic non-stationary patterns from the input sequences. In the second stage, the full model is trained end-to-end, with the MPPM parameters either fixed or jointly updated, while the Forecasting Model (FM) optimizes the upper-level forecasting objective. This decoupled approach stabilizes training, mitigates error propagation, and enhances overall forecasting performance. By leveraging the high-quality amplitude and phase representations learned in the first stage, TimeAPN produces accurate and robust predictions. Furthermore, this separation preserves the framework’s model-agnostic nature, enabling adaptation to diverse forecasting scenarios without architectural modifications.

The lower-level optimization is performed using the Adaptive Moment Estimation (ADAM) optimizer. This effectively decomposes the original non-stationary forecasting problem into two relatively independent tasks: amplitude and phase-difference prediction, and the backbone model’s forecasting. For the lower-level task, mean squared error (MSE) is used to quantify the discrepancy between predicted and ground-truth amplitude and phase sequences. The upper-level objective is similarly optimized using ADAM, with MSE measuring the difference between the final predictions and the ground-truth future series.
Formally, the training objective can be expressed as:
\begin{equation}
\label{eq:original_bilevel}
\begin{aligned}
\arg \min_{\theta} \; & 
\sum_{(\bm{y}^c, \bm{y}^c_*)} 
\mathcal{L}_{\mathrm{FM}}\!\left(\theta, \phi^{*}, (\bm{y}^c, \bm{y}^c_*)\right), \\
\text{s.t.} \quad 
\phi^{*} = \arg \min_{\phi} \; &
\beta\sum_{\substack{\bm{\mu},\bm{w}}} 
\mathcal{L}_{\mathrm{MPPM}}\!\left(
\theta, \phi, 
\bm{\mu},\bm{w}\right) \nonumber \\
& \quad + 
\left(1-\beta\right)\sum_{\bm{A}} 
\mathcal{L}_{\mathrm{MPPM}}\!\left(
\theta, \phi, 
\bm{A}
\right).
\end{aligned}
\end{equation}
where
\begin{equation*}
\begin{aligned}
\bm{\mu}  &= (\bm{\mu}_y^c, \Delta\bm{\mu}^c),\
\bm{w} &= (\bm{w}^c_y, \Delta\bar{\bm{w}}^c),\
\bm{A} &= (\bm{A}^c_y, \Delta\bar{\bm{A}}^c).
\end{aligned}
\end{equation*}
Here, $\theta$ and $\phi$ denote the parameters of the FM and MPPM, respectively, and $\beta$ is a learnable weight. $\bm{y}^c$ represents the ground-truth future series for channel $c$, while $\bm{\mu}^c_y$, $\bm{w}^c_y$, and $\bm{A}^c_y$ are the mean, phase, and amplitude of $\bm{y}^c$, and $\Delta\bar{\bm{A}}^c$ is the amplitude of the predicted series $\bm{y}^c*$. Integrating amplitude information alongside mean and phase information ensures that the model effectively captures rapid energy changes, thereby improving prediction performance.

\begin{table*}[ht]
\centering
\small
\caption{
Comparison of TimeAPN with existing reversible normalization methods under different prediction lengths (96, 192, 336, and 720). Results are reported in terms of MSE using two backbone models, FEDformer and DLinear, across multiple benchmark datasets. IMP denotes the relative improvement achieved by TimeAPN over the original backbone without normalization. The best results for each setting are highlighted in \textbf{bold}.}
\renewcommand{\arraystretch}{1.06}
\setlength{\tabcolsep}{4pt}

\resizebox{\textwidth}{!}{%
\begin{tabular}{cc|
ccccccc|ccccc}
\toprule

\multicolumn{2}{c}{\multirow{2}{*}{Methods}} &
\multicolumn{7}{c|}{FEDformer} &
\multicolumn{5}{c}{DLinear} \\

\multicolumn{1}{c}{} &
\multicolumn{1}{c|}{} &
\multicolumn{1}{c}{+TimeAPN} & 
\multicolumn{1}{c}{+DDN} & 
\multicolumn{1}{c}{+RevIN} &
\multicolumn{1}{c}{+NST} &
\multicolumn{1}{c}{+Dish-TS} &
\multicolumn{1}{c}{+SAN} &
\multicolumn{1}{c|}{IMP} &
\multicolumn{1}{c}{+TimeAPN} & 
\multicolumn{1}{c}{+DDN} & 
\multicolumn{1}{c}{+RevIN} &
\multicolumn{1}{c}{+SAN} &
\multicolumn{1}{c}{IMP} \\
\bottomrule

\multirow{4}{*}{\rotatebox{90}{ETTh1}} 
       & 96  & \textbf{0.377} & 0.389 & 0.396 & 0.394 & 0.390 & 0.384 & -0.5\% & \textbf{0.371} & 0.372 & \textbf{0.371} &  0.381 & 3.4\% \\
       & 192 & \textbf{0.424} & 0.425 & 0.439 & 0.441 & 0.441 & 0.430 & 0.7\% & \textbf{0.405} & 0.405 & 0.406 &  0.419 & 8.6\%\\
       & 336 & \textbf{0.466} & 0.474 & 0.484 & 0.485 & 0.495 & 0.474 & -1.7\% & \textbf{0.430} & 0.431 & 0.442 &  0.438 & 3.8\%\\
       & 720 & \textbf{0.498} & 0.679 & 0.515 & 0.505 & 0.519 & 0.504 & 54.4\% & \textbf{0.435} & 0.457 & 0.442 &  0.446 & 13.7\%\\
\midrule

\multirow{4}{*}{\rotatebox{90}{ETTh2}} 
        & 96  & \textbf{0.288} & 0.299 & 0.385 & 0.381 & 0.806 & 0.302 & 15.5\% & \textbf{0.272} & 0.278 & 0.278 & 0.277 & 6.2\% \\
        & 192 & \textbf{0.377} & 0.391 & 0.484 & 0.478 & 0.936 & 0.403 & 12.9\% & \textbf{0.334} & 0.342 & 0.336 & 0.341 & 13.9\% \\
        & 336 & 0.432 & \textbf{0.422} & 0.560 & 0.561 & 1.039 & 0.490 & 14.1\% & \textbf{0.360} & 0.364 & 0.362 & 0.361 & 22.2\% \\
        & 720 & 0.439 & \textbf{0.437} & 0.486 & 0.502 & 1.237 & 0.465 & 8.5\% & 0.398 & 0.396 & \textbf{0.391} & 0.398 & 45.7\%\\
\midrule

\multirow{4}{*}{\rotatebox{90}{ETTm1}} 
        & 96  & \textbf{0.308} & 0.341 & 0.338 & 0.336 & 0.348 & 0.310 & 15.2\% & \textbf{0.287} & 0.288 & 0.303 & 0.289 & 4.7\%\\
        & 192 & \textbf{0.350} & 0.382 & 0.386 & 0.386 & 0.406 & 0.349 & 12.3\% & \textbf{0.325} & 0.326 & 0.338 & 0.324 & 3.3\%\\
        & 336 & \textbf{0.391} & 0.424 & 0.441 & 0.438 & 0.438 & 0.396 & 11.1\% & 0.358 & \textbf{0.357} & 0.373 & \textbf{0.357} & 3.8\%\\
        & 720 & \textbf{0.443} & 0.477 & 0.491 & 0.483 & 0.497 & 0.442 & 10.0\% & 0.413 & 0.417 & 0.431 & \textbf{0.410} & 3.3\%\\
\midrule

\multirow{4}{*}{\rotatebox{90}{ETTm2}} 
        & 96  & \textbf{0.168} & 0.173 & 0.191 & 0.191 & 0.394 & 0.176 & 11.6\% & 0.168 & 0.169 & \textbf{0.165} & 0.166 & 2.3\%\\
        & 192 & \textbf{0.236} & 0.239 & 0.262 & 0.270 & 0.552 & 0.247 & 7.8\% & 0.223 & 0.226 & \textbf{0.220} & 0.221 & 5.9\%\\
        & 336 & \textbf{0.302} & 0.310 & 0.342 & 0.353 & 0.808 & 0.325 & 7.4\% & 0.275 & 0.282 & 0.273 & \textbf{0.268} & 6.8\%\\
        & 720 & \textbf{0.400} & 0.414 & 0.444 & 0.445 & 1.282 & 0.415 & 8.9\% & 0.363 & 0.373 & 0.367 & \textbf{0.357} & 15.0\%\\
\midrule

\multirow{4}{*}{\rotatebox{90}{Weather}} 
        & 96  & \textbf{0.173} & 0.220 & 0.187 & 0.187 & 0.244 & 0.177 & 27.9\% & \textbf{0.146} & 0.150 & 0.175 & 0.151 & 16.1\%\\
        & 192 & \textbf{0.228} & 0.291 & 0.235 & 0.235 & 0.320 & 0.243 & 49.2\% & \textbf{0.190} & 0.194 & 0.218 & 0.197 & 12.8\%\\
        & 336 & \textbf{0.285} & 0.349 & 0.288 & 0.289 & 0.424 & 0.321 & 20.4\% & \textbf{0.239} & 0.244 & 0.266 & 0.246 & 9.1\%\\
        & 720 & \textbf{0.356} & 0.404 & 0.361 & 0.359 & 0.604 & 0.381 & 13.4\% & \textbf{0.309} & 0.314 & 0.334 & 0.315 & 6.9\%\\
\midrule

\multirow{4}{*}{\rotatebox{90}{Electricity}} 
        & 96  & \textbf{0.148} & 0.146 & 0.171 & 0.172 & 0.175 & 0.163 & 23.4\% & \textbf{0.128} & 0.131 & 0.141 & 0.137 & 8.6\%\\
        & 192 & \textbf{0.160} & 0.162 & 0.184 & 0.187 & 0.188 & 0.179 & 17.8\% & \textbf{0.147} & 0.148 & 0.155 & 0.152 & 4.5\%\\
        & 336 & 0.177 & \textbf{0.172} & 0.204 & 0.202 & 0.209 & 0.193 & 16.9\% & \textbf{0.160} & 0.165 & 0.171 & 0.167 & 5.3\%\\
        & 720 & 0.198 & \textbf{0.197} & 0.229 & 0.230 & 0.239 & 0.221 & 18.2\% & \textbf{0.187} & 0.205 & 0.210 & 0.202 & 8.3\%\\
\midrule

\multirow{4}{*}{\rotatebox{90}{Traffic}} 
        & 96  & 0.449 & \textbf{0.443} & 0.617 & 0.612 & 0.613 & 0.532 & 21.9\% & \textbf{0.374} & 0.375 & 0.412 & 0.410 & 9.4\%\\
        & 192 & 0.470 & \textbf{0.465} & 0.639 & 0.641 & 0.644 & 0.563 & 23.3\% & \textbf{0.393} & 0.396 & 0.424 & 0.429 & 7.3\%\\
        & 336 & \textbf{0.476} & \textbf{0.476} & 0.657 & 0.654 & 0.659 & 0.574 & 23.3\% & \textbf{0.409} & 0.411 & 0.437 & 0.447 & 4.9\%\\
        & 720 & 0.520 & \textbf{0.515} & 0.682 & 0.688 & 0.693 & 0.605 & 17.5\% & \textbf{0.447} & 0.452 & 0.465 & 0.473 & 4.1\%\\
\bottomrule
\end{tabular}%
} 
\label{tab:longterm1}
\end{table*}

\section{Experiments}
In this section, we evaluate the performance of the TimeAPN method against advanced deep forecasters.

\textbf{Datasets \& Baselines.}
We conduct extensive experiments on seven widely used real-world datasets, including the Electricity Transformer Temperature (ETT) benchmark with four subsets (ETTh1, ETTh2, ETTm1, and ETTm2), as well as Weather, Electricity, and Traffic. These datasets are commonly used to evaluate long-term time-series forecasting models and are publicly available under the benchmark protocol in \cite{wu2021autoformer}.
TimeAPN is a model-agnostic framework that can be integrated into various time-series forecasting architectures. To demonstrate its general applicability, we incorporate TimeAPN into several representative models, including FEDformer~\cite{zhou2022fedformer}, DLinear~\cite{zeng2023transformers}, PatchTST~\cite{nie2022time}, and S\textunderscore Mamba~\cite{wang2025mamba}, which cover diverse architectural paradigms in modern forecasting methods.

 
\textbf{Implementation Details.}
The Mean Square Error (MSE) and Mean Absolute Error (MAE) are chosen as evaluation metrics, with MSE serving as the training loss. All models use the same prediction lengths $T = \{96, 192, 336, 720\}$. For the look-back window $L$, FEDformer, PatchTST and  S\textunderscore Mamba use $L = 96$, while DLinear utilizes $L = 336$. For the Traffic dataset, PatchTST and S\textunderscore Mamba instead adopt a larger look-back window of $L=720$. The wavelet bases are initialized to the biorthogonal 3.5 wavelet bases. $\alpha$ and $\beta$ start at one.

\subsection{Main Results}
Results reported in Table~\ref{tab:longterm} show that the integration of TimeAPN consistently enhances the forecasting accuracy of all four baseline models across benchmark datasets. The performance gains are especially pronounced under the MSE metric on larger-scale datasets, including Weather, Electricity, and Traffic. Specifically, when equipped with TimeAPN, FEDformer achieves relative error reductions of $28.6\%$, $18.92\%$, and $20.83\%$ on these datasets, respectively, while DLinear attains corresponding reductions of $10.44\%$, $6.75\%$, and $6.78\%$. Comparable improvements are also observed when TimeAPN is incorporated into the remaining baseline models. It is worth noting that FEDformer, DLinear, and PatchTST do not adopt reversible normalization in their official implementations, whereas S\textunderscore Mamba applies the RevIN technique based on fixed statistical properties. Even so, substituting the RevIN module in S\textunderscore Mamba with the proposed TimeAPN module still leads to clear performance improvements. Overall, these empirical findings convincingly indicate that TimeAPN enhances the effectiveness of baseline models for TSF tasks.

\subsection{Comparison With Reversible Normalization Methods}
\noindent{\textbf{Quantitative Evaluation.}}
We compare representative reversible normalization methods using the MSE metric across four prediction lengths {96, 192, 336, 720}. As shown in Table~\ref{tab:longterm1}, integrating TimeAPN consistently improves the forecasting performance of the backbone models and achieves superior results compared with existing reversible normalization approaches, including RevIN~\cite{kim2021reversible}, NST~\cite{liu2022non}, Dish-TS~\cite{fan2023dish}, SAN~\cite{liu2023adaptive}, and DDN~\cite{dai2024ddn}.
NST is specifically designed for Transformer-based architectures and cannot be applied to DLinear due to the absence of attention mechanisms. We also evaluated Dish-TS with DLinear; however, it consistently degraded performance across datasets, and thus its results are omitted from the table.
Overall, TimeAPN outperforms DDN and other normalization methods in most settings, demonstrating strong robustness across datasets and prediction horizons. The improvements are particularly notable on larger datasets such as Weather and Electricity, as well as on the ETT benchmarks (ETTh1 and ETTh2), indicating that TimeAPN is more effective at handling complex non-stationary patterns in long-term time-series forecasting.


Considering that early models often lacked robust generalizability due to their naive modeling strategies, we additionally included comparisons with two recent representative models: PatchTST and S\textunderscore Mamba. These methods already exhibit good non-stationary adaptability, so they can better reflect the upper performance limit of fine-grained normalization methods. As illustrated in Table~\ref{tab:longterm2}, TimeAPN significantly outperforms the recent state-of-the-art model DDN in handling non-stationary information. Overall, TimeAPN achieves the best performance compared to DDN across all forecasting cases. Specifically, SAN and DDN exhibit unstable performance in long-term forecasting settings, with limited performance or even worse than the original predictive model across multiple datasets and prediction lengths. Conversely, TimeAPN consistently improves forecasting accuracy across all backbones, including FEDformer, DLinear, PatchTST, and S\textunderscore Mamba. These results indicate that modeling non-stationarity at finer temporal granularity enables TimeAPN to better handle long-term distributional shifts, yielding more reliable and robust forecasting performance.
\begin{table*}[ht]
\centering
\small
\caption{Comparison of forecasting errors for DDN, SAN, and TimeAPN. The best results are highlighted in \textbf{bold}.}

\resizebox{0.9\textwidth}{!}{%
\begin{tabular}{cc|
cccccc|cccccc}
\toprule

\multicolumn{2}{c}{\multirow{2}{*}{Methods}} &
\multicolumn{6}{c}{PatchTST} &
\multicolumn{6}{c}{S\textunderscore Mamba} \\

\multicolumn{1}{c}{} &
\multicolumn{1}{c}{} &
\multicolumn{2}{c}{+TimeAPN} & 
\multicolumn{2}{c}{+DDN} & 
\multicolumn{2}{c}{+SAN} &
\multicolumn{2}{c}{+TimeAPN} & 
\multicolumn{2}{c}{+DDN} & 
\multicolumn{2}{c}{+SAN} \\
\cmidrule(lr){1-2}\cmidrule(lr){3-4}\cmidrule(lr){5-6}\cmidrule(lr){7-8}\cmidrule(lr){9-10}\cmidrule(lr){11-12}\cmidrule(lr){13-14}
\multicolumn{2}{c}{Metric} & MSE & MAE & MSE & MAE & MSE & MAE & MSE & MAE & MSE & MAE & MSE & MAE \\
\bottomrule

\multirow{4}{*}{\rotatebox{90}{ETTh2}} 
        & 96  & \textbf{0.289} & \textbf{0.338} & 0.291 & 0.341 & 0.304 & 0.358 & \textbf{0.290} & \textbf{0.340} & 0.291 & 0.343 & 0.316 & 0.368 \\
        & 192 & \textbf{0.369} & \textbf{0.390} & 0.378 & 0.392 & 0.417 & 0.434 & \textbf{0.372} & \textbf{0.394} & 0.383 & 0.402 & 0.400 & 0.416 \\
        & 336 & \textbf{0.409} & \textbf{0.426} & 0.418 & 0.428 & 0.427 & 0.446 & \textbf{0.413} & \textbf{0.430} & 0.429 & 0.436 & 0.418 & 0.434 \\
        & 720 & \textbf{0.402} & \textbf{0.431} & 0.430 & 0.446 & 0.428 & 0.455 & \textbf{0.420} & \textbf{0.443} & 0.428 & 0.444 & 0.447 & 0.463 \\
\midrule


\multirow{4}{*}{\rotatebox{90}{ETTm2}} 
        & 96  & \textbf{0.175} & \textbf{0.257} & 0.182 & 0.268 & 0.194 & 0.290 & \textbf{0.174} & \textbf{0.259} & 0.177 & 0.264 & 0.182 & 0.270 \\
        & 192 & \textbf{0.241} & \textbf{0.301} & 0.248 & 0.312 & 0.257 & 0.331 & \textbf{0.241} & \textbf{0.304} & 0.245 & 0.311 & 0.246 & 0.318 \\
        & 336 & \textbf{0.301} & \textbf{0.340} & 0.307 & 0.342 & 0.349 & 0.391 & \textbf{0.301} & \textbf{0.341} & 0.304 & 0.349 & 0.307 & 0.361 \\
        & 720 & \textbf{0.397} & \textbf{0.397} & 0.407 & 0.399 & 0.435 & 0.444 & \textbf{0.396} & \textbf{0.401} & 0.402 & \textbf{0.401} & \textbf{0.396} & 0.413 \\
\midrule

\multirow{4}{*}{\rotatebox{90}{Weather}} 
        & 96  & \textbf{0.167} & \textbf{0.219} & 0.177 & 0.224 & 0.172 & 0.235 & \textbf{0.155} & \textbf{0.220} & 0.166 & 0.225 & 0.171 & 0.231 \\
        & 192 & \textbf{0.214} & \textbf{0.264} & 0.218 & 0.267 & 0.212 & 0.273 & \textbf{0.205} & \textbf{0.264} & 0.210 & 0.271 & 0.213 & 0.271 \\
        & 336 & 0.266 & \textbf{0.305} & \textbf{0.265} & 0.314 & 0.262 & 0.313 & \textbf{0.254} & \textbf{0.303} & 0.271 & 0.322 & 0.263 & 0.309 \\
        & 720 & \textbf{0.332} & \textbf{0.358} & 0.333 & 0.368 & \textbf{0.332} & 0.360 & \textbf{0.332} & 0.360 & 0.346 & 0.380 & 0.336 & \textbf{0.361} \\
\midrule

\multirow{4}{*}{\rotatebox{90}{Electricity}} 
        & 96  & \textbf{0.151} & \textbf{0.252} & 0.174 & 0.265 & 0.171 & 0.262 & \textbf{0.135} & \textbf{0.240} & 0.142 & 0.243 & 0.145 & 0.245 \\
        & 192 & \textbf{0.171} & \textbf{0.268} & 0.180 & 0.273 & 0.177 & 0.270 & \textbf{0.154} & \textbf{0.259} & 0.159 & 0.262 & 0.159 & 0.260 \\
        & 336 & \textbf{0.189} & 0.287 & 0.195 & 0.290 & 0.191 & \textbf{0.285} & \textbf{0.170} & \textbf{0.276} & 0.175 & 0.280 & 0.179 & 0.282 \\
        & 720 & \textbf{0.224} & \textbf{0.317} & 0.233 & 0.325 & 0.225 & \textbf{0.317} & \textbf{0.196} & 0.304 & 0.208 & 0.312 & 0.201 & \textbf{0.302} \\

\bottomrule
\end{tabular}%
} 
\label{tab:longterm2}
\end{table*}

\begin{figure*}[ht]
\centering
\includegraphics[width=1\textwidth]{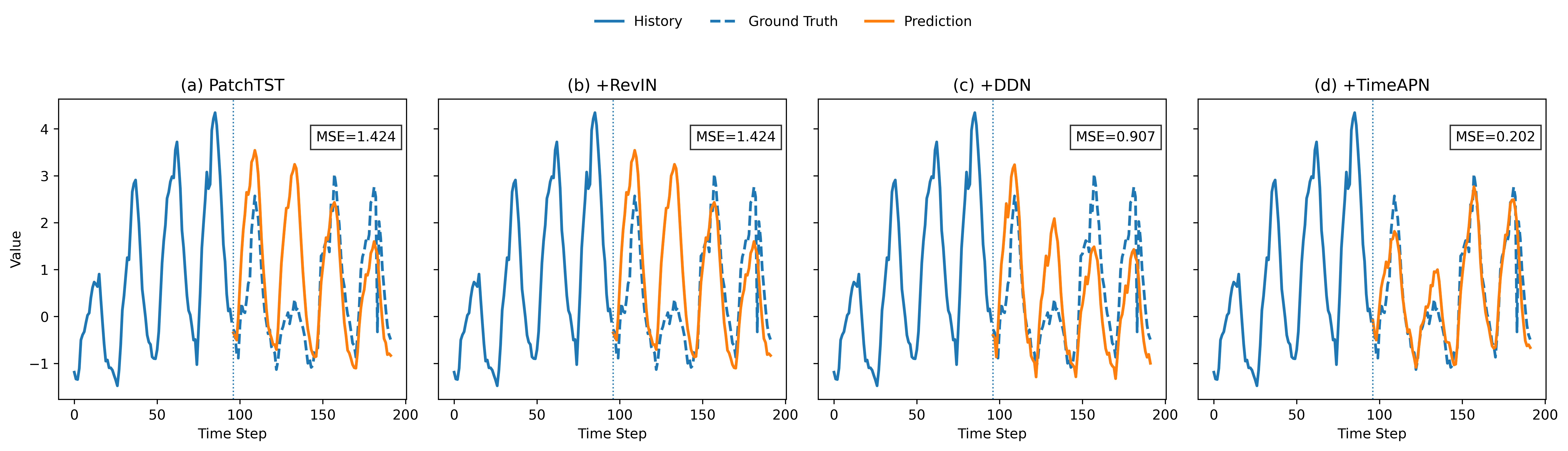}
\caption{
Comparison of reversible normalization methods on a forecasting sample from the PatchTST model trained on the Electricity dataset. (a) PatchTST without normalization, (b) PatchTST with RevIN, (c) PatchTST with DDN, and (d) PatchTST with the proposed TimeAPN.}
\label{03_fig}
\end{figure*}

\begin{table*}[ht]
\centering
\small
\caption{Impact of amplitude, phase, and frequency-domain components on forecasting performance. The best results are highlighted in \textbf{bold}.}

\resizebox{0.7\textwidth}{!}{%
\begin{tabular}{c|
ccccc|ccccc}
\toprule

\multicolumn{1}{c|}{\multirow{2}{*}{Methods}} &
\multicolumn{5}{c|}{PatchTST} &
\multicolumn{5}{c}{S\textunderscore Mamba} \\

\multicolumn{1}{c|}{} &
\multicolumn{1}{c}{Amp} & 
\multicolumn{1}{c}{Phase} & 
\multicolumn{1}{c}{FD} &
\multicolumn{1}{c}{MSE} &
\multicolumn{1}{c|}{MAE} &
\multicolumn{1}{c}{Amp} &
\multicolumn{1}{c}{Phase} &
\multicolumn{1}{c}{FD} & 
\multicolumn{1}{c}{MSE} & 
\multicolumn{1}{c}{MAE} \\
\bottomrule



\multirow{4}{*}{\rotatebox{90}{ETTm1}} 
       & $\times$ & $\times$ & $\times$ & 0.382 & \textbf{0.395} & $\times$ & $\times$ & $\times$ & 0.398 & 0.407 \\
       & $\times$  & $\checkmark$  & $\checkmark$  & 0.379 & 0.397 & $\times$  & $\checkmark$  & $\checkmark$  & 0.392 & 0.405  \\
       & $\checkmark$ & $\checkmark$ & $\times$ & 0.386 & 0.400 & $\checkmark$ & $\checkmark$ & $\times$ & 0.396 & 0.405 \\
       & $\checkmark$ & $\checkmark$ & $\checkmark$ & \textbf{0.377} & \textbf{0.395} & $\checkmark$ & $\checkmark$ & $\checkmark$ & \textbf{0.391} & \textbf{0.404} \\
\midrule


\multirow{4}{*}{\rotatebox{90}{Weather}} 
        & $\times$ & $\times$ & $\times$ & 0.263 & 0.283 & $\times$ & $\times$ & $\times$ & 0.251 & \textbf{0.276} \\
       & $\times$  & $\checkmark$  & $\checkmark$  & \textbf{0.243} & 0.289 & $\times$  & $\checkmark$  & $\checkmark$  & 0.239 & 0.288  \\
       & $\checkmark$ & $\checkmark$ & $\times$ & 0.258 & 0.292 & $\checkmark$ & $\checkmark$ & $\times$ & 0.258 & 0.300 \\
       & $\checkmark$ & $\checkmark$ & $\checkmark$ & 0.245 & \textbf{0.287} & $\checkmark$ & $\checkmark$ & $\checkmark$ & \textbf{0.237} & 0.287 \\
\midrule

\multirow{4}{*}{\rotatebox{90}{Electricity}} 
        & $\times$ & $\times$ & $\times$ & 0.207 & 0.289 & $\times$ & $\times$ & $\times$ & 0.171 & 0.268 \\
       & $\times$  & $\checkmark$  & $\checkmark$  & 0.185 & 0.282 & $\times$  & $\checkmark$  & $\checkmark$  & 0.170 & 0.274  \\
       & $\checkmark$ & $\checkmark$ & $\times$ & 0.189 & 0.282 & $\checkmark$ & $\checkmark$ & $\times$ & 0.170 & 0.273 \\
       & $\checkmark$ & $\checkmark$ & $\checkmark$ & \textbf{0.184} & \textbf{0.281} & $\checkmark$ & $\checkmark$ & $\checkmark$ & \textbf{0.164} & \textbf{0.270} \\

\bottomrule
\end{tabular}%
} 
\label{tab:analysis}
\end{table*}

\noindent{\textbf{Qualitative Evaluation.}}
Fig.~\ref{03_fig} compares the proposed TimeAPN with existing reversible normalization approaches, including RevIN and DDN, highlighting their different mechanisms for addressing time-series non-stationarity.
In subplots (a) and (b), PatchTST without normalization and with RevIN are presented. RevIN reconstructs the predicted sequence using the statistics of the historical window, thereby aligning the forecast distribution with the historical distribution. While effective when the future distribution is similar to the past, this assumption may break down under distribution shifts, resulting in degraded forecasting performance.
Subplot (c) shows the results of DDN, which mitigates non-stationarity through normalization in both the time and frequency domains. By applying a sliding-window strategy, DDN captures variations in frequency-wise distributions across temporal segments. However, its normalization relies on fixed-period statistics and primarily models first- and second-order moments, which may limit its ability to capture rapid and fine-grained temporal fluctuations.
In contrast, subplot (d) demonstrates the behavior of the proposed TimeAPN. TimeAPN performs adaptive normalization by jointly modeling amplitude and phase variations using both time- and frequency-domain representations. Its point-wise adaptive mechanism enables dynamic adjustment to changes in local distributions, allowing the model to better capture fine-grained temporal dynamics. As a result, TimeAPN produces forecasts that more accurately follow the underlying signal patterns, leading to improved forecasting accuracy in scenarios with rapid or irregular fluctuations.

\subsection{Analysis}

To better understand the contribution of different frequency-domain components, we conduct an ablation study by selectively enabling the amplitude (Amp), phase (Phase), and frequency decomposition (FD) modules. The results are summarized in Table IV. Overall, each component contributes positively to forecasting performance, and their joint integration consistently produces the best results across datasets.

When all frequency-related components are removed, the model exhibits a noticeable performance decline, suggesting that relying solely on time-domain modeling is insufficient for addressing the non-stationarity inherent in long-term forecasting tasks. Introducing phase modeling significantly improves prediction accuracy, indicating that phase information is important for capturing temporal alignment and structural dependencies in time-series data. Incorporating amplitude modeling further enhances performance by enabling the model to better represent variations in signal magnitude and intensity.

Moreover, combining Amp and Phase already yields substantial improvements over using either component alone, highlighting their complementary roles in frequency-domain modeling. When FD is additionally incorporated, the model achieves the best overall performance on both the ETTm2 and Weather datasets under the PatchTST and S\textunderscore Mamba backbones. This result suggests that explicit frequency decomposition helps disentangle temporal patterns with different periodicities and dynamics, enabling more precise modeling of complex time-series structures.

Overall, these findings demonstrate that jointly modeling amplitude, phase, and frequency components leads to more expressive and robust representations of non-stationary time series, thereby improving multivariate long-term forecasting performance. The consistent gains across different backbone architectures further validate the generality and effectiveness of the proposed frequency-domain design.

\section{Conclusion}

This paper presents TimeAPN, an adaptive normalization framework for multivariate long-term time-series forecasting that explicitly addresses non-stationarity. Unlike existing normalization approaches that rely primarily on mean and variance statistics, TimeAPN models amplitude and phase dynamics to capture rapid and complex distributional changes commonly observed in real-world time series. By jointly exploiting time-domain and frequency-domain representations, the proposed framework provides a more expressive and fine-grained characterization of temporal dynamics.

TimeAPN is designed as a model-agnostic, plug-and-play module that can be seamlessly integrated into diverse forecasting architectures via a reversible normalization–denormalization process. Extensive experiments on multiple benchmark datasets demonstrate that incorporating TimeAPN consistently improves forecasting accuracy across backbone models and prediction horizons, and outperforms existing reversible normalization methods, particularly under challenging non-stationary conditions.

Overall, this work highlights the importance of explicit amplitude–phase modeling for robust time-series forecasting and suggests a promising direction for handling complex non-stationary dynamics. 

\bibliographystyle{IEEEtran}   
\bibliography{references} 

\end{document}